# A Unified Framework for Human–AI Collaboration in Security Operations Centers with Trusted Autonomy

AHMAD MOHSIN, Centre for Securing Digital Futures, School of Science, Edith Cowan University, Australia
HELGE JANICKE, Centre for Securing Digital Futures, School of Science, Edith Cowan University, Australia
AHMED IBRAHIM, Centre for Securing Digital Futures, School of Science, Edith Cowan University, Australia
IQBAL H. SARKER, Centre for Securing Digital Futures, School of Science, Edith Cowan University, Australia
SEYIT CAMTEPE, CSIRO's Data61, Australia

This article presents a structured framework for Human-AI collaboration in Security Operations Centers (SOCs), integrating AI autonomy, trust calibration, and Human-in-the-loop decision making. Existing frameworks in SOCs often focus narrowly on automation, lacking systematic structures to manage human oversight, trust calibration, and scalable autonomy with AI. Many assume static or binary autonomy settings, failing to account for the varied complexity, criticality, and risk across SOC tasks considering Humans and AI collaboration. To address these limitations, we propose a novel autonomy tiered framework grounded in five levels of AI autonomy from manual to fully autonomous, mapped to Human-in-the-Loop (HITL) roles and task-specific trust thresholds. This enables adaptive and explainable AI integration across core SOC functions, including monitoring, protection, threat detection, alert triage, and incident response. The proposed framework differentiates itself from previous research by creating formal connections between autonomy, trust, and HITL across various SOC levels, which allows for adaptive task distribution according to operational complexity and associated risks. The framework is exemplified through a simulated cyber range that features the cybersecurity AI-Avatar, a fine-tuned LLM-based SOC assistant. The AI-Avatar case study illustrates human-AI collaboration for SOC tasks, reducing alert fatigue, enhancing response coordination, and strategically calibrating trust. This research systematically presents both the theoretical and practical aspects and feasibility of designing next-generation cognitive SOCs that leverage AI not to replace but to enhance human decision-making.

CCS Concepts: • **Security and privacy → Security operations**; • **Computing methodologies → Artificial intelligence**; *Cognitive science.*

Additional Key Words and Phrases: Human–AI Collaboration, Security Operations Centers (SOCs), AI Autonomy and Trust, Cognitive SOC Framework, Cybersecurity Automation

Authors' Contact Information: Ahmad Mohsin, a.mohsin@ecu.edu.au, Centre for Securing Digital Futures, School of Science, Edith Cowan University, Perth, Western Australia, Australia; Helge Janicke, h.janicke@ecu.edu.au, Centre for Securing Digital Futures, School of Science, Edith Cowan University, Perth, Western Australia, Australia; Ahmed Ibrahim, ahmed.ibrahim@ecu.edu.au, Centre for Securing Digital Futures, School of Science, Edith Cowan University, Perth, Western Australia, Australia; Iqbal H. Sarker, i.sarker@ecu.edu.au, Centre for Securing Digital Futures, School of Science, Edith Cowan University, Perth, Western Australia, Australia; Seyit Camtepe, seyit.camtepe@data61.csiro.au, CSIRO's Data61, Sydney, New South Wales, Australia.

## 1 Introduction

A Security Operations Center (SOC) serves as a centralized unit in cybersecurity, bringing together personnel, processes, and technology to protect an organization's digital assets. As described in cybersecurity literature, an SOC functions as a proactive defensive hub, constantly monitoring, detecting, analyzing, and responding to cyber threats in real time within IT, operational technology (OT), and cloud settings. The SOC acts as the control center for cybersecurity tasks, integrating threat intelligence, security monitoring, incident response, and regulatory compliance to address emerging cyber risks. By utilizing advanced security analytics, automation, and machine learning, SOCs can detect anomalies, correlate security incidents, and mitigate threats before they develop into widespread breaches [21, 52]. Their importance is heightened in sectors at high risk, such as finance, healthcare, critical infrastructure, and government, where even a minor security incident can lead to dire operational and financial impacts. As cyber threats grow more advanced, organizations are increasingly depending on SOCs to bolster their security measures, reduce risks, and ensure continuous business operations amid an evolving threat environment.

Amid the surge of state-sponsored cyberattacks and financially driven cybercriminal activities, SOCs have emerged as the primary defense against advanced and evolving threats. In 2024, there has been a 44% increase in cyberattacks worldwide, with key targets being critical infrastructure organizations, particularly from state actors. Emerging cyber threats such as cyber espionage and disruptive incursions into IT/OT are now major strategic concerns [8]. The threat environment continues to evolve, with vulnerabilities rising by up to 17%. By 2027, cybercrime is forecasted to cause damages amounting to $23 trillion [49], underscoring the essential need for strong cybersecurity operations. As adversaries adopt more sophisticated AI tools and techniques, organizations are increasingly committing resources to modernize SOCs, integrating automation, AI-powered analytics, and orchestration to boost efficiency. With global spending on cybersecurity expected to hit $500 billion by 2030 [15], there is an intensified drive to strengthen SOC capabilities, especially through AI-based automation and analytics, to address these advancing cyber threats.

Despite the existing automation and AI-driven SOC capabilities, cybersecurity teams face overwhelming challenges in defending systems against cyber threats. A major issue is alert fatigue. SOCs generate thousands of security alerts daily, many of which are false positives. A global study found that SOC teams encounter 4,484 alerts per day, leading to 67% of alerts being ignored due to analysts being overloaded with many tasks. Studies show that up to 99% of SOC alerts can be false positives [9, 11, 19], contributing to analyst overload and 'alert fatigue'. Indeed, analysts often struggle with high false-positive rates, cognitive fatigue, and a shortage of skilled personnel [5].

Another challenge is the rapid progression of attacks, often outpacing human response times. Adversaries launch attacks faster than analysts can react, overwhelming them with complex threat detection and investigation tasks, highlighting the need for AI-driven automation. Although SOCs are equipped with diverse security technologies, including Security Information and Event Management (SIEM) systems, their effectiveness is often limited by extensive tuning requirements and high volumes of uncorrelated alerts [9]. Most SIEM platforms heavily rely on rule-based correlation and analytics, offering some automation but requiring significant human intervention, making cybersecurity operations largely manual and reactive [32, 42]. Security Orchestration, Automation, and Response (SOAR) platforms aim to streamline workflows, but rigid, predefined rules often fail to adapt to novel attack patterns in real time. Additionally, Endpoint Detection and Response (EDR) and Extended Detection and Response (XDR) solutions enhance endpoint and cross-domain visibility, but they struggle with alert overload and lack adaptive threat detection against evolving attack techniques.

To address these challenges, organizations are increasingly integrating Artificial Intelligence (AI) and automation to enhance SOC operations. AI-driven security analytics can correlate data across multiple attack surfaces, automate incident response, and provide predictive threat intelligence. AI modelling techniques, i.e., Machine Learning (ML) /Deep Learning (DL), improve threat detection by identifying behavioral anomalies that deviate from baseline activities, thereby reducing false positives. Additionally, AI-powered automation enables SOC teams to prioritize, investigate, and contain threats more efficiently than traditional manual workflows [5, 9, 11]. A significant advancement in this domain is the application of Natural Language Processing (NLP) techniques, which enhance cybersecurity workflows by assisting in breach protection, threat identification, and scope analysis. The development of Large Language Models (LLMs), such as OpenAI's ChatGPT, Google's Gemini, and Meta's open-source LLaMA models, has further revolutionized SOC operations [20]. LLMs have the potential to process unstructured security data, analyze logs, summarize incidents, and assist in real-time decision-making. These LLMs enable natural language-based interactions, allowing analysts to query security intelligence more efficiently and collaborate with AI-driven assistants for incident response. By leveraging NLP and LLM advancements, SOCs can enhance their capabilities to detect, analyze, and respond to threats in a more streamlined and effective manner. Despite significant advancements in AI-driven SOC automation, existing efforts lack a comprehensive vision for evaluating Human-AI collaboration across the full spectrum of cybersecurity operations. Current approaches often overlook the balance between automation and human oversight, failing to integrate Human-in-the-Loop (HITL) strategies that ensure effective decision-making. We present a Human-AI Collaboration Framework that focuses on augmentation, teaming, and adaptive automation for SOC operations.

This article presents an overview of the present landscape of AI applications in cybersecurity, exploring the trends in SOC automation, the capabilities of AI, and the obstacles faced in collaboration between humans and AI. The SOC functions are methodically aligned with the NIST-based incident response lifecycle [41], highlighting the essential activities within a SOC. This work offers significant contributions by formalizing a triadic model of autonomy, trust, and HITL collaboration, offering a comprehensive conceptual foundation that unifies theoretical modelling with practical implementation. Effective Human–AI collaboration in SOCs requires an approach in which autonomy levels are dynamically aligned with the complexity of SOC functions, the criticality of business operations, and the degree of trust required for decision-making. The proposed five-level autonomy framework directly addresses these challenges by mapping autonomy levels to SOC analyst roles, ranging from semi-autonomous (AI-augmented) configurations with HITL to fully autonomous settings, based on task complexity, thereby enhancing scalability and operational clarity. Various Human-AI collaboration components of the framework are exemplified for the SOC incident response life cycle, using an AI-Avatar fine-tuned in a simulated cyber range, and models real-world SOC dynamics for Human-AI collaboration. Developed under the ACDC project, the AI-Avatar, a refined ChatGPT-based LLM with RAG and knowledge graphs, operates in a simulated SOC with SIEM, IDS, and firewalls. Trained over two years with wargaming, event data, and Red/Blue Team exercises, it aids analysts in threat detection, triage, and response. The results indicate significant advancements in SOC efficiency, a decrease in analyst workload, and enhanced trust-based Human-AI decision-making. We engage in a critical examination and analysis of optimizing AI-driven autonomous and augmented SOC operations for future developments.

The remainder of this paper is organized as follows: Section 2 reviews SOC operations, AI evolution, and Human–AI collaboration challenges. Section 3 introduces the foundational concepts of our framework. Section 4 presents the Human–AI Collaboration Framework, focusing on autonomy, trust, and task allocation. Related work and comparative analysis are provided in Section 5. Section 6 illustrates Human–AI collaboration through a case study. Section 7 discusses key contributions and limitations. Section 8 outlines future directions and concludes the paper.

### 1.1 SOCs Capabilities and Security Operations

Technological advancements and the growing sophistication of cyber threats have driven the continuous evolution of SOCs. As adversaries increasingly employ advanced and evasive tactics, the need for rapid, accurate, and scalable threat detection and response has made the integration of AI and automation essential in modern SOC workflows.

*Early SOCs and Traditional SIEM Systems.* Initial SOC implementations heavily relied on Security Information and Event Management (SIEM) systems that aggregated and normalized event logs from various sources to identify suspicious activities [54]. These systems typically employed rule-based correlation, signature-based detection, and keyword pattern matching to detect known threats such as brute-force attempts or malware signatures [3, 33]. However, such static detection mechanisms struggled with adaptability, resulting in high false-positive rates and ineffective detection of zero-day threats and Advanced Persistent Threats (APTs) [52]. Statistical anomaly detection methods were also introduced to identify deviations from behavioral baselines, but lacked the adaptive learning required to distinguish benign anomalies from true threats, further contributing to alert fatigue [3]. While SIEMs provided centralized visibility, inconsistencies in log formats and limitations in cross-platform correlation hindered their effectiveness. As a result, early SOCs remained heavily dependent on manual triage and investigation, delaying response times and burdening analysts with repetitive tasks.

*Integrating SOAR, EDR, and XDR Technologies.* To enhance operational agility and reduce manual workload, SOCs began integrating Security Orchestration, Automation, and Response (SOAR) platforms, Endpoint Detection and Response (EDR) tools, and more recently, Extended Detection and Response (XDR) solutions [16]. SOAR platforms introduced playbook-driven automation, executing predefined workflows in response to routine security events such as phishing, malware detection, or privilege misuse [23]. These playbooks use conditional logic to trigger contextual actions, streamlining response processes and reducing analyst intervention. EDR tools enhance endpoint visibility by monitoring activities, detecting abnormal behavior, and automating containment. XDR expands this by integrating telemetry across various environments for unified threat detection [16]. SOCs use API-driven orchestration for seamless integration of SIEMs, firewalls, EDR, ticketing systems, and threat intelligence platforms. This allows for automated actions like blocking IPs, isolating endpoints, revoking access, and generating reports with minimal input. By correlating multi-source data, incident enrichment improves context, quickens investigations, and reduces alert fatigue. Despite progress, human oversight is needed for novel threats, highlighting the demand for AI systems with explainability and human-in-the-loop features to balance automation and accountability.

## 2 Cybersecurity Operations and Challenges in SOCs

Despite advancements in cybersecurity technologies, Security Operations Centers (SOCs) continue to face challenges. These challenges stem from (i) functional/operational issues and (ii) internal organizational barriers within SOC capabilities, which are elaborated on in the subsections below.

### 2.1 SOC Functional Challenges

We present the operational hurdles faced by SOCs, correlating them with the NIST Cybersecurity Framework (CSF) [41], a widely recognized industry benchmark for the incident response lifecycle. By mapping and analyzing SOC functional challenges in the context of the incident response lifecycle, we provide practical insight into the challenges faced by SOCs, as elaborated below:

**1. Asset Management.** Effective asset identification is vital to SOC performance, yet SIEM and EDR tools have

Table 1. SOC functions with descriptions, contributing failure factors, and operational impacts, aligned with NIST CSF phases. Highlights key systemic weaknesses, including technological gaps and Human–AI limitations impacting SOC effectiveness.

| SOC Function | Description | Contributing Factors | SOC Operations Impact |
|---|---|---|---|
| **Assets Management (Identify)** | | | |
| **Asset Discovery** | Accurate inventory maintenance is difficult. Incomplete logs reduce visibility of lateral movement. | • SIEM/EDR gaps<br>• Log misconfigurations<br>• Aggregation challenges | • Expanded attack-surface<br>• Missed indicators<br>• Delayed response |
| **System Hardening and Protection (Protect)** | | | |
| **Monitor & Protect** | Static or misconfigured controls create exploitable blind spots. | • Outdated rules<br>• Inadequate training<br>• Lack of adaptability | • Unauthorized access<br>• Data breaches<br>• Persistent threats |
| **Threat Management (Detect)** | | | |
| **Detection & Mitigation** | False positives and analyst skill gaps weaken threat management. | • Alert overload<br>• Escalation delays<br>• Analyst limitations<br>• Weak AI tools | • Escalated incidents<br>• Remediation cost<br>• Threat persistence |
| **Alert Management (Detect/Respond)** | | | |
| **Triage & Validation** | Excessive alerts exhaust analysts, risking missed threats. | • Static scoring<br>• Analyst burnout<br>• Partial knowledge-bases<br>• Weak SOAR automation | • Missed threats<br>• Operational delay<br>• Escalated attacks |
| **Incident Response and Recovery (Respond/Recover)** | | | |
| **Response & Recovery** | Poor coordination and manual processes delay response. | • Manual workflows<br>• Patch delays<br>• Tier misalignment | • Downtime<br>• Compliance risk<br>• Higher costs |

limitations in real-time asset discovery, especially in complex IT/OT environments. Manual processes and the transient nature of complex interdependent systems result in incomplete inventories, leaving endpoints undamaged. The largest portion of effort and resources is dedicated to collecting and aggregating security data management. Inaccurate or missing logs, particularly from domain controllers, can obscure lateral movement by attackers, undermining detection efforts [54]. The absence of automated discovery and continuous asset validation increases vulnerabilities and impairs situational awareness, tying reliable log management to asset management. This is compounded by log capture errors and inadequate storage, posing security threats by expanding the attack surface.

**2. Monitoring & Protection.** Protection serves as a crucial function within a SOC, aiming to avert unauthorized system and data access. Its efficiency is frequently compromised by inflexible security configurations, inadequate upkeep, and insufficient employee training. Improperly set up firewalls or IDS may either overlook sophisticated threats or produce excessive false positives, resulting in security gaps [24]. Adversaries like APTs exploit vulnerabilities through lateral movement or zero-day exploits. To counter this, SOCs are using automation and machine learning. Tools like adaptive firewalls, anomaly-based intrusion prevention, and behavior analytics respond to new threats, reducing manual work and boosting data security. They can detect subtle activities like gradual data exfiltration that traditional methods miss. However, enhanced protection requires advanced technology, skilled personnel, and AI support.

**3. Threat Management.** A typical SOC employs SIEMs, EDR tools, and anomaly detection to find threats. However, these systems often generate false alarms, tiring analysts, and sometimes miss real threats. AI lacks context for recognizing subtle attacks or correlating lateral movement clues. Analyst and tool limitations, complex processes, and skepticism toward automation can delay mitigation [24, 54]. Effective threat management needs precise detection, thorough investigations, and quick response. SIEMs and EDRs overwhelm analysts with false alerts, complicating threat identification. AI has difficulty adapting to new attacks, leaving threats unresolved. Analyst skills, tool limitations, slow decisions, and distrust of automation impede swift mitigation. AI's lack of contextual understanding and inability to deduce intentions lead to oversight. Analysts may overlook stealthy malware and traffic anomalies due to SIEM's failure in correlating EDR logs with network data [21]. Threat management efficiency is further impacted by a shortage of skilled personnel and cognitive overload. Too many alerts cause fatigue, diminishing analysts' focus and accuracy in threat evaluation.

**4. Alerts Management.** SOC alert pipelines are inundated with alerts, with as much as 80% proving to be false positives owing to improperly configured detection rules and subpar models. This necessitates manual filtering by analysts, which postpones detection and response efforts, leading to fatigue and burnout [54, 58]. The absence of SOAR integration obstructs efficient prioritization and the correlation of data across various tools and resources. Alerts are meant to address cyber threats, but are often plagued by tool inadequacies, human errors, and operational challenges, leading to fatigue and delays [3]. Alerts stem from security and endpoint tools and go through analysis to differentiate threats from anomalies. Effective SOC management requires optimizing detection rules and automating responses with AI-driven prioritization and SOAR systems to reduce false positives. Enhancing SOC involves automation, agile intelligence, Human-AI collaboration, and real-time correlation to minimize noise and manual work.

**5. Incident Response and Recovery.** Effective incident response and recovery are critical for minimizing damage and restoring normal operations after a security breach. However, SOCs face significant challenges that impede this incident response and recovery [12]. Organizations encounter difficulties in incident detection and escalation due to inadequate monitoring, which prolongs exposure to vulnerabilities. According to the 2024 IBM Cost of a Data Breach Report, organizations took an average of 204 days to identify a breach and an additional 73 days to contain it, totaling 277 days duration that underscores the critical need for timely detection mechanisms [18]. Manual ticketing and isolated task management delay incident response due to inefficient workflows and poor coordination. Ineffective communication can prolong breach containment, worsening threats. High-severity incidents like ransomware attacks require urgency, but outdated tools and organizational issues often delay response. Automated playbooks can lower recovery times, but doubts about AI reliability and compatibility with legacy systems hinder adoption. Legacy systems rely on outdated tech, complicating AI integration and posing security risks.

### 2.2 Human, Organizational and Technological Challenges

From a human perspective, SOCs face ongoing challenges of ***operational overload, alert fatigue, and stress.*** Analysts struggle with massive alert volumes, 90% being false positives [3, 18, 58], causing constant triage and neglect of true positives. Fatigue from irrelevant alerts increases the risk of missing real threats [18]. Stress from data overload and lack of ***unified tools*** further weakens SOC performance, burying critical signals and affecting morale, leading to high turnover. With over 40% of SOCs ***understaffed*** [30], fewer analysts handle complex scenarios [54], facing skill gaps due to insufficient training and rapid threat evolution. Burnout and attrition are rising due to long hours, low autonomy, and lack of recognition, resulting in higher errors and slower responses [3, 58]. Without support like clear escalation paths or flexible schedules, SOC performance suffers. AI-powered automation offers potential relief, improving detection and response times. However, issues with explainability, adversarial vulnerability, and integration with old systems limit its use. Trust in AI requires transparency and collaboration. Despite technical capabilities, organizational barriers like legacy policies and departmental silos delay crucial responses, particularly during significant incidents.

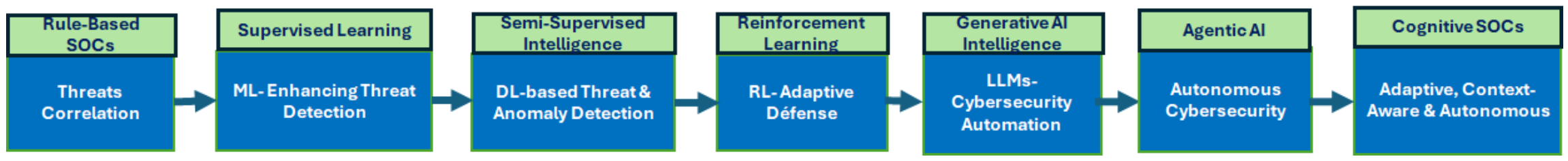

Fig. 1. AI Evolution for Cybersecurity in SOCs.

### 2.3 SOC Functions Failures and Impacts

Based on the challenges identified above, we now systematically provide an analysis of SOC function failures, their contributing factors, and the impacts on cybersecurity operations. The effectiveness of SOCs largely depends on executing essential cybersecurity tasks, including asset management, system defense, threat monitoring, alert handling, and incident response, which are often compromised by operational challenges and interrelated barriers. Through this analysis, we aim to highlight the critical importance of understanding these failure points, as they directly influence SOC resilience and operational stability. We investigate how each failure scenario weakens system security, identifying key factors contributing to SOC failures. Besides operational issues, privacy concerns, and diverse data aggregation complicate decision-making, increasing misclassification risks. Table 1 outlines SOC functions and related failures in this review, highlighting vulnerabilities and informing strategies to improve SOC performance and strengthen cybersecurity.

To ensure alignment with recognized security standards, each SOC functional category in the Table 1 is systematically mapped to the relevant phases of the NIST Cybersecurity Framework[1] (Identify, Protect, Detect, Respond, Recover) [41], enabling a structured understanding of operational gaps and their implications. Failures in SOC functions, such as ineffective asset identification, alert fatigue, and delayed incident response, undermine cybersecurity resilience and operational efficiency. High false-positive rates in threat detection lead to cognitive fatigue and alert dismissal by analysts [51]. Inadequate AI integration limits adaptive threat identification and incident response, creating vulnerabilities [35]. Adaptive AI-driven SOC frameworks dynamically balance human oversight and automation, improving threat detection, scalability, and operational resilience while minimizing analyst workload and decision fatigue [29, 40].

### 2.4 AI Evolution in Cybersecurity and SOCs

AI has evolved from rule-based systems in early SIEM and SOC platforms to advanced paradigms such as Machine Learning (ML), Deep Learning (DL), Generative AI, and more recently, Agentic AI [42, 44]. Figure 1 illustrates AI's expanding role in enhancing threat detection, classification, and response. As cyber threats grow in complexity, organizations are increasingly adopting AI-driven strategies to strengthen cybersecurity [21, 59]. These techniques are applied across the SOC lifecycle, including monitoring, detection, response, and recovery, each offering distinct advantages and operational challenges [42, 59].

**Traditional SOC Rule-Based Systems.** Before the emergence of data-driven AI techniques, SOCs primarily depended on rule-based systems for threat detection. These systems operated using predefined signatures, heuristics, and manually configured correlation rules to identify known attack patterns. While effective against well-understood threats, they lacked the flexibility to detect novel or zero-day attacks, as they could not learn or adapt from emerging

[1]While NIST CSF 2.0 includes a dedicated "Governance" function, governance aspects from a Human–AI collaboration perspective within SOCs are considered out of scope for this article

threat data [32]. The reliance on manual rule creation also contributed to high false positive rates, increasing analyst workload and alert fatigue, which in turn delayed response times. These constraints highlight the inflexibility of static detection techniques, which have prompted the progression of AI in SOCs from machine learning to deep learning and the adaptive, AI-driven strategies discussed below.

**Supervised Learning (ML): Enhancing Threat Detection.** These are mostly ML architectures. Supervised learning models leverage labeled datasets to classify cyber threats, making them effective for tasks such as intrusion detection, malware classification, and spam filtering. Common techniques include Random Forests (RF), Support Vector Machines (SVM), Decision Trees (DT), and Deep Neural Networks (DNNs), a foundational approach in deep learning. For example, one study applied RF and Gradient Boosting Machines (GBM) in SOC-based intrusion detection, achieving high accuracy for known threats; however, performance degraded against novel attack patterns, underscoring the limitations of labeled-data dependence [21]. Similarly, hybrid models like CNN-LSTM, combining convolutional feature extraction with sequence learning, have shown promise in malware detection but require frequent retraining to remain resilient against evolving threats [2]. Supervised models classify known threats efficiently in real-time but struggle with zero-day attacks, high labeling costs, model drift, and adversarial manipulation. They also face data imbalance and high false positives, especially with new or dynamic threats.

**Semi-Supervised Learning (DL): Threats & Anomaly Detection.** Semi-supervised learning, often combined with deep learning architectures, is frequently employed in cybersecurity. This method utilizes a limited amount of labeled data together with a much larger set of unlabeled data, making it particularly beneficial in situations where labeled threat information is limited. Prominent techniques include Graph Neural Networks (GCNs) and Variational Autoencoders (VAEs). A study introduced a GNN framework combining GCNs with a semi-supervised autoencoder [36], trained on both labeled and unlabeled data, improving threat detection recall by 18% over unsupervised methods [38, 59]. Methods combining a VAE with an SVM classifier reduce false positives by extracting latent features before classification [36, 38]. These models are pretrained on unlabeled data through autoencoders, contrastive learning, or clustering, then fine-tuned with labeled data using standard loss functions. Semi-supervised learning enhances detection in low-label environments and identifies subtle anomalies signaling emerging security threats.

**Reinforcement Learning (RL): Adaptive Defense Mechanisms.** Reinforcement Learning models learn by interacting with the environment, which makes them valuable for developing adaptive cyber defense strategies. RL approaches such as Deep Q-Networks (DQN) models enable SOC agents to learn optimal responses through trial and error. Studies have shown that RL agents can outperform traditional rule-based intrusion detection systems, though they require extensive training and high-quality historical data [21, 39]. RL models typically employ Markov Decision Processes (MDPs) to maximize a security-based reward function and are trained using algorithms like Policy Gradient Methods and Proximal Policy Optimization (PPO). In practice, RL supports:

- *Automated Incident Response*–> Learning optimal responses to various attack scenarios to reduce response times.
- *Dynamic Resource Allocation*–> Efficiently allocating defensive resources based on the current threat landscape.

Despite their promise, RL models face challenges in transparency (often functioning as “black boxes”), require substantial computational resources, and present integration challenges in real-world SOC environments.

**Cybersecurity Automation with Generative AI.** Generative AI, powered by advanced deep learning architectures such as Long Short-Term Memory (LSTM) networks and Transformers, is revolutionizing cybersecurity. By generating synthetic data, it enriches training datasets and improves the accuracy of threat detection models. Additionally, Generative AI supports secure software development and vulnerability discovery, reinforcing organizational resilience [20, 34].

Deep learning and generative architectures, including LSTM networks, model temporal dependencies in network traffic, while Transformer-based models further enhance detection accuracy in cloud and enterprise environments [4, 6, 26]. Building on these advances, such models also support knowledge graph construction via Large Language Models (LLMs), enabling enriched threat intelligence and contextual analysis. Transformer-based LLMs like BERT and GPT extend these capabilities by being fine-tuned for tasks such as threat intelligence extraction and anomaly detection [28]. Practical implementations include Google's Sec-PaLM [10], which incorporates Mandiant and VirusTotal data, and Microsoft's Security Copilot [31], which integrates with SIEM and XDR tools to enhance investigation, threat hunting, and automated SOC response.

**Agentic AI.** Agentic AI, grounded in Large Language Models (LLMs), Chain-of-Thought (CoT) reasoning, and coordinated Agentic Workflows, enables a shift from reactive to proactive cybersecurity operations in future SOCs [20, 25, 34]. These systems autonomously analyze large-scale telemetry and threat intelligence data to generate real-time, scalable insights. CoT enhances transparency in decision-making and enables agents to perform complex tasks, such as identifying IOCs and assessing incident impact [22]. Agentic Workflows orchestrate task-specific agents across the incident response lifecycle, from detection to recovery, while LLM-driven agents proactively forecast vulnerabilities and recommend actionable mitigation strategies.

**Towards Cognitive Security Operation Centers.** The evolution of AI is fundamentally transforming cybersecurity, shifting from static, rule-based tools to adaptive, context-aware systems that actively support threat detection, anomaly analysis, and incident response. Modern SOCs increasingly employ natural language processing (NLP), machine learning/deep learning (ML/DL), and generative AI across large language models (LLMs) and multi-agent frameworks. These advancements have given rise to *Cognitive SOCs*, where AI assists analysts by correlating threat data, triaging incidents, and automating repetitive tasks. Such systems enhance detection accuracy, reduce response time, and increase operational resilience. However, early SOC automation remained largely reactive, requiring manual rule-tuning and lacking the adaptability needed to counter sophisticated threats such as zero-day exploits, Advanced Persistent Threats (APTs), and AI-driven attacks. These limitations, along with persistent challenges in scalability, interpretability, and human expertise, highlight the need for structured Human–AI collaboration models. By balancing autonomy with oversight, such models can improve trust, ensure contextual understanding, and enable more proactive, resilient cybersecurity operations.

Building on the evolution of AI capabilities in SOC environments and in response to both functional and organizational challenges, the following section introduces foundational concepts essential for effective Human–AI collaboration, which are central to the design and operation of Cognitive SOCs.

## 3 Human-AI Collaboration in SOCs

Human–AI collaboration presents unique challenges in high-stakes domains such as cybersecurity, where decision-making requires both rapid response and contextual judgment in fast-paced and uncertain changing environments. While AI offers scalability and speed [9, 50], it often lacks explainability and situational awareness, key traits of human cognition. Achieving synergy between AI autonomy and human oversight is difficult due to conflicting requirements: trust must be earned through consistent performance, while excessive automation risks alienating human analysts. This tension makes it challenging to harmonize autonomy, trust, and collaboration. Human–AI collaboration in SOCs leverages complementary strengths through three core pillars: Autonomy, Human-in-the-Loop (HITL), and Trust (Figure 2).

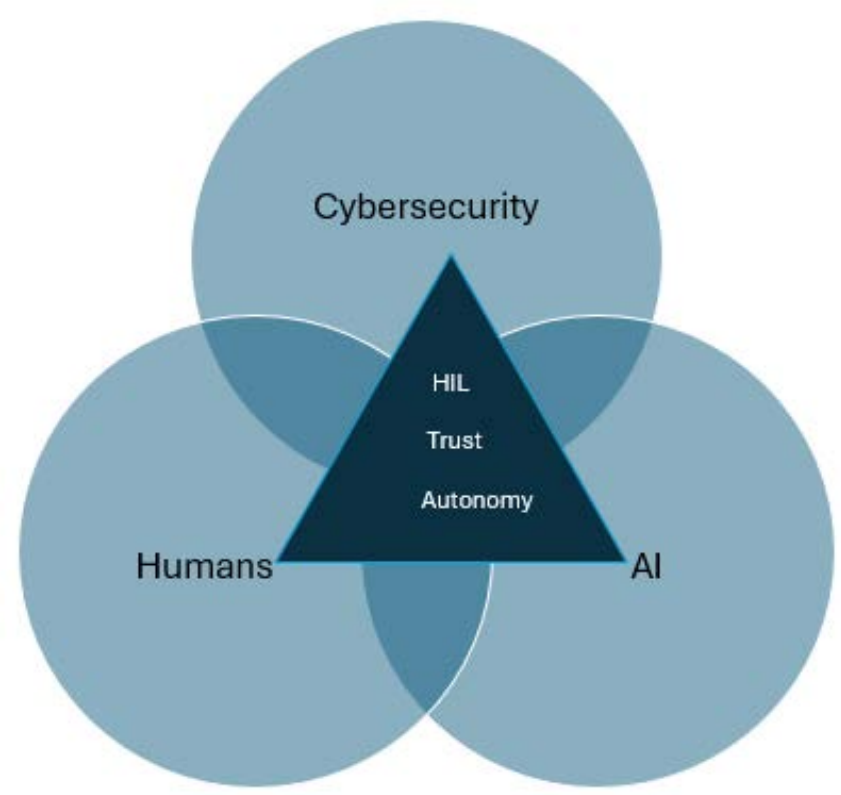

Fig. 2. Core Pillars of Human–AI Collaboration in SOCs: The triadic relationship between Autonomy (A), Human-in-the-Loop (HITL), and Trust (T). Higher trust enables scalable autonomy while reducing human intervention. Balancing these dimensions ensures efficient, explainable, and accountable cybersecurity operations.

### 3.1 Autonomy

AI autonomy refers to a system's ability to independently execute tasks and make decisions without continuous human intervention [27, 56]. In the context of SOCs, increasing levels of autonomy can significantly improve efficiency and response times. AI can react to incidents in milliseconds, far faster than human operators [60]. Autonomy in SOC operations spans a spectrum from fully supervised (low autonomy) to fully independent execution (high autonomy). The optimal level balances efficiency with risks like automation complacency and reduced contextual awareness [1, 53].

We define autonomy as a scalar $A \in [0, 1]$, modeled as:

$$A = 1 - (\lambda_1 C + \lambda_2 R)(1 - T) \tag{1}$$

Task complexity $C$, risk $R$, and trust $T$ are normalized in $[0, 1]$ with weighting factors $\lambda_1, \lambda_2 \in [0, 1]$. Higher complexity and risk reduce autonomy, whereas increased trust enhances it. SOC tasks vary greatly in complexity and risk, impacting the required cognitive effort and contextual awareness, and thus influence the suitable AI autonomy level. Trust $T$ moderates this, facilitating safe AI delegation.

### 3.2 Human-in-the-Loop

The Human-in-the-Loop (HITL) paradigm integrates human judgment into AI-driven workflows, which is vital during uncertain, novel, or high-stakes incidents [5, 29]. As autonomy increases, human involvement tends to decrease [35], forming a critical inverse relationship for balanced and accountable decision-making.

We define HITL involvement $H \in [0, 1]$ as:

$$H = 1 - A \tag{2}$$

Here, $H = 1$ reflects full human control (zero autonomy), while $H = 0$ indicates complete AI autonomy. Effective HITL designs help prevent automation bias, maintain situational awareness, and ensure ethical oversight [29, 35].

### 3.3 Trust

Trust reflects an analyst's confidence in AI's transparency, competence, and reliability. Properly calibrated trust prevents both overtrust, which can lead to neglect, and undertrust, which limits AI utility [1, 46, 57].

Trust $T \in [0, 1]$ is influenced by explainability $E$, performance history $P$, and uncertainty $U$, with:

$$T = \alpha_1 E + \alpha_2 P + \alpha_3 (1 - U) \tag{3}$$

where $\alpha_1 + \alpha_2 + \alpha_3 = 1$. Greater transparency, reliable past performance, and low uncertainty contribute to higher trust, which in turn enables higher autonomy and reduced human oversight, enhancing SOC agility and efficiency.

### 3.4 Human SOC Role Tiers

The responsibilities of Security Operations Center (SOC) tiers can be systematically aligned with the functional domains defined in the NIST Cybersecurity Framework (CSF) 2.0 [41]. This alignment enables a structured and proactive approach to cybersecurity operations by mapping specific analyst roles to key cybersecurity functions. Tiered structuring not only improves operational efficiency but also clarifies the interaction between human analysts and AI-driven systems in SOC workflows [37, 55].

- *Tier 1* primarily supports the *Identify* and *Protect* functions. Analysts at this level are responsible for system inventory, vulnerability identification, baseline configuration, and enforcement of preventive controls. They perform real-time monitoring, initial alert triage, and escalate incidents as needed.
- *Tier 2* aligns with the *Detect* and *Respond* functions, involving detailed threat investigation, correlation of security events, and the execution of containment strategies. Tier 2 analysts often act as incident handlers, coordinating with other teams to manage security breaches.
- *Tier 3* spans the *Detect* and *Recover* domains. These senior analysts fine-tune detection mechanisms, lead threat hunting activities, conduct malware and forensics investigations, and assist in the development of recovery procedures.
- *Tier 4* corresponds to the *Govern* and *Recover* functions. Responsibilities include policy oversight, risk management, compliance enforcement, and high-level decision-making during complex incidents such as ransomware outbreaks or critical infrastructure attacks. This tier is often engaged in strategic planning and aligning SOC goals with enterprise cybersecurity governance

We establish SOC human analyst roles that not only clarify responsibilities across SOC tiers but also provide a foundation for integrating AI systems into SOC workflows. This facilitates the design of tier-aware AI autonomy models that support Human-in-the-Loop (HITL) decision-making, enabling effective automation while preserving human oversight in critical security operations.

The following section presents a Human–AI Collaboration Framework for SOCs, building upon the foundational concepts and operational structures outlined above. This framework addresses key challenges in SOC operations and aims to improve decision-making efficiency, enhance situational awareness, and strengthen the overall resilience of modern security operations centers.

## 4 Human-AI Collaboration Framework for Security Operations Centers

This section presents a structured Human–AI Collaboration Framework tailored for Security Operations Centers (SOCs), addressing the challenges outlined in Section 2 and highlighting AI's growing significance in cybersecurity tasks.

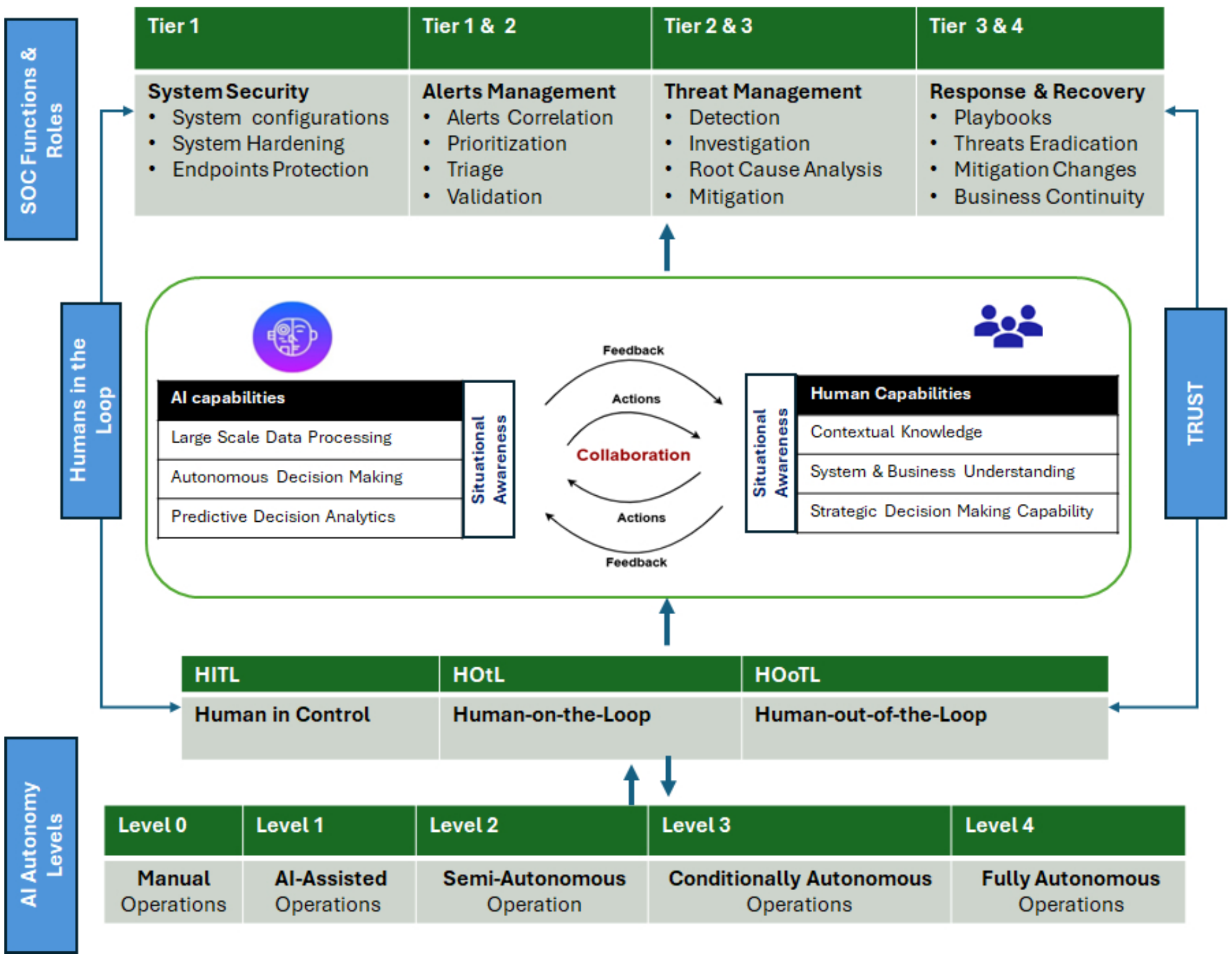

Fig. 3. Human-AI Collaboration Framework in SOCs with a layered mapping SOC Functional Tiers to increasing levels of AI autonomy *(A)* (Level 0 to Level 4), aligned with human roles ranging from Human-in-the-Loop (HITL) to Human-out-of-the-Loop (HOoTL). At the core, a bidirectional collaboration loop between AI and human capabilities mediated by situational awareness and trust enables adaptive decision-making, feedback-driven learning, and scalable automation in cybersecurity operations.

Conventional SOCs frequently face hurdles such as alert fatigue, false positives, slow response times, and analyst exhaustion. While advancements in ML/DL models, SIEM/SOAR systems, and LLM-based tools have enhanced threat detection and operational automation, human expertise remains essential for understanding intricate threats, making critical decisions, and ensuring ethical practices. Current research suggests that AI can positively impact SOC challenges and its role in cybersecurity. Consequently, we propose a Human-AI Collaboration Framework to integrate AI with human analysts, with foundational elements discussed in Section 3. SOCs face alert fatigue, false positives, and human burnout, lacking a full incident response cycle for threat management and decision-making. Effective AI integration needs a structured strategy. Our framework offers a multi-layered method for merging AI with human collaboration in SOCs.

Figure 3 illustrates the proposed framework. The bottom layer defines five Autonomy levels (Levels 0–4), aligned with HITL, Human-on-the-Loop (HOtL), and Human-out-of-the-Loop (HOoTL) roles. As trust in AI systems grows, autonomy can be scaled incrementally, ensuring that governance, accountability, and operational safety remain intact. This layered structure supports flexible, modular deployment across diverse SOC environments. From the top layer, this framework organizes SOC functions across four tiers System Security, Alerts Management, Threat Management, and Response & Recovery mapped to increasing levels of AI autonomy and corresponding human oversight roles. Routine, high-volume tasks are increasingly delegated to AI agents (Levels 1–3), while analysts retain strategic control over critical decisions. At the core of the framework is a symbiotic interaction between human and AI capabilities. AI contributes large-scale data processing, predictive analytics, and autonomous decision-making, while humans provide contextual reasoning, business awareness, and ethical judgment. Feedback loops and transparency mechanisms (e.g., explainability panels, uncertainty scoring) support trust calibration and autonomy scaling.

### 4.1 Autonomy for AI Driven SOC

Autonomy, as defined earlier in Section 3, describes an AI system's ability to independently perform tasks and make informed decisions without continuous human intervention. Within SOCs, managing autonomy effectively is crucial to improving response times and operational efficiency while maintaining essential human oversight. This careful balance prevents risks like automation complacency or inappropriate autonomous decisions [1, 48, 56].

Previously, we defined autonomy in 1 as:

$$A = 1 - (\lambda_1 C + \lambda_2 R)(1 - T)$$

where $A$ is autonomy, $C$ is task complexity, $R$ is risk, and $T$ is trust, each normalized between 0 (low) and 1 (high), and weights $\lambda_1, \lambda_2 \in [0, 1]$. We systematically develop autonomy scale categorization ranges for AI autonomy levels in SOC operations, as described in Table 2. This scale is inspired by the SAE J3016 standard [43], which defines six levels of driving automation for on-road motor vehicles, ranging from Level 0 (no automation) to Level 5 (full automation) [45] . By adapting these principles, we structure SOC autonomy into five clear operational levels, illustrated below with practical examples involving common SOC technologies like SIEM, SOAR, and AI agents:

Table 2. AI Autonomy Levels in SOCs and Corresponding Ranges

| **Level 0 Autonomy** | **Level 1 Autonomy** | **Level 2 Autonomy** | **Level 3 Autonomy** | **Level 4 Autonomy** |
|---|---|---|---|---|
| $A \approx 0$ | $A \approx 0.2$–$0.4$ | $A \approx 0.4$–$0.6$ | $A \approx 0.7$–$0.8$ | $A \approx 0.9$–$1.0$ |

**Level 0 – Manual Operations (No Autonomy).** Analysts manually handle all security alerts and responses without automation. *Example: Analysts manually review raw logs in SIEM systems, manually correlate firewall events, and execute incident response tasks (like blocking IPs or isolating endpoints) individually, causing delays and increased human error potential.*

**Level 1 – AI-Assisted Operations (Decision Support).** AI assists by providing alert prioritization or recommending investigative actions through SIEM dashboards or ML models. Analysts maintain decision control. *Example: A SIEM employs machine learning to score alerts by severity. Analysts manually review high-severity alerts and approve recommended actions suggested by the system, such as manually resetting passwords or quarantining suspicious files.*

**Level 2 – Semi-Autonomous Operations (AI Acts with Human Approval).** AI systems such as SOAR tools automate routine detections and propose response workflows, awaiting explicit analyst approval for critical actions. *Example: A SOAR platform automatically triggers endpoint isolation workflows upon detecting malware but awaits analyst confirmation before isolating critical servers or privileged accounts. Analysts receive concise notifications and quickly approve or reject these recommended actions.*

**Level 3 – Conditionally Autonomous Operations (Human-in-the-Loop).** Advanced AI agents [25] autonomously execute defined security tasks, involving analysts only in exceptional scenarios. Analysts monitor these systems in a supervisory manner (human-on-the-loop). *Example: AI-driven SOAR and security analytics tools autonomously detect phishing emails and remove them from mailboxes, block malicious IP addresses, and isolate compromised endpoints. Analysts supervise these autonomous processes through periodic review or intervene only when AI flags uncertainties or novel attack patterns.*

**Level 4 – Fully Autonomous Operations (Minimal Human Oversight).** Highly autonomous AI agents manage comprehensive SOC functions with minimal human involvement. Analysts' roles shift primarily to governance, audit, and policy management. *Example: An advanced AI-driven SOC platform automatically ingests threat intelligence, proactively hunts threats across enterprise environments, autonomously contains sophisticated ransomware incidents by isolating endpoints, initiates recovery procedures, generates incident reports, and continuously updates its own detection models. Analysts periodically audit AI decisions and provide strategic oversight without intervening in routine operational tasks.*

In the subsection below, we integrate a key component of our proposed framework: the role of Humans-in-the-Loop (HIL) in conjunction with SOC AI autonomy levels, emphasizing the balance between trust and autonomy.

### 4.2 SOCs AI: Human-in-the-Loop

The Human-in-the-Loop (HITL) paradigm embeds human judgment into AI-driven workflows, which is especially critical during ambiguous, novel, or high-risk cyber incidents [5]. We leverage the HITL concept introduced in Section 3, defining it formally as inversely proportional to autonomy: higher HITL levels correspond to lower autonomy, and vice versa, following HIL description in 2 we have:

$$H = 1 - A$$

where $H \in [0, 1]$ represents the degree of human control, and $A \in [0, 1]$ denotes the level of autonomy. At $H = 1$, humans retain full control; at $H = 0$, AI operates autonomously.

**Levels of Human–AI Involvement in SOCs.**

We define three levels of HITL to capture varying degrees of human involvement across the SOC autonomy spectrum:

(1) *Human-in-Control (full HITL).* Humans make all operational decisions; AI provides support (e.g., scoring alerts or suggesting actions). Common at *Autonomy Level 1*, as seen in SIEM-driven SOCs where analysts manually investigate and respond.
(2) *Human-on-the-Loop (HOtL).* AI executes predefined tasks under conditional human oversight. Analysts monitor operations and intervene when alerts require contextual judgment. Typical of *Autonomy Levels 2–3*, such as SOAR playbooks that automate response with analyst escalation for complex cases.
(3) *Human-out-of-the-Loop (HOoTL).* AI autonomously manages routine tasks with minimal human interaction, primarily for audit or policy verification. Found in *Autonomy Level 4* settings where AI handles end-to-end threat response (e.g., malware isolation, access revocation).

Relying too much on full HITL slows operations and hinders AI efficiency *automation bottlenecks*. On the other hand, full HOoTL settings may result in AI actions without human awareness, leading to unnoticed misclassifications or new threats *loss of oversight*. Effective HITL needs *trust and training*; analysts should understand AI, oversee decisions, and intervene when needed [29, 35].*Tier-1 analysts* work in HITL or HOtL modes, handling alerts and initial responses with AI aid. *Tier-2 analysts* are Human-on-the-Loop, managing workflows and AI decisions. *Tier-3 analysts*, at the top, operate in a Human-out-of-the-Loop role, focusing on threat hunting, AI oversight, and ensuring ethics and policy adherence. Dynamic HITL levels, adjusted for task complexity, risk, and AI trust, combined with autonomy calibration (Section 4.1) and trust management (Section 3.3), form a foundation for secure Human-AI collaboration in SOCs.

### 4.3 Trust: Confidence pillar between Humans and AI

Trust represents the third foundational pillar of the Human–AI Collaboration Framework, mediating the relationship between autonomy and human-in-the-loop involvement. As autonomy increases, direct human oversight diminishes, making calibrated trust essential for safely delegating operational decisions to AI systems in SOCs. Formally, trust $T$ is modelled as a function of explainability, performance history, and uncertainty (see Equation 3), encapsulating the key factors that influence analysts' confidence in AI systems. At lower autonomy levels (Levels0–1), trust has minimal operational effect, as human analysts maintain complete control and manually validate AI outputs. In this setting, AI typically operates in a supportive role, offering recommendations such as alert prioritization or threat scoring. Importantly, trust and autonomy are interlinked: as trust increases, the system can safely assume greater autonomy (Equation1), while high task complexity and risk constrain it. This correlation reinforces the need for calibrated trust to unlock operational scalability without compromising safety.

As SOCs evolve toward intermediate autonomy (Levels2–3), where AI executes predefined actions under human oversight (e.g., SOAR playbooks), trust becomes operationally significant. Overtrust can result in automation bias and oversight of errors, while undertrust introduces unnecessary redundancy and reduces efficiency[1, 57]. To support calibrated trust, our framework emphasizes the role of *Explainable AI (XAI)*. Systems must provide interpretable justifications for their actions, such as highlighting key features or indicators behind flagged alerts, to facilitate rapid analyst validation, particularly under high-alert volume conditions [1, 51].

Challenges remain as black-box models limit interpretability and adversarial inputs undermine trust [1, 53]. Trust in AI needs continuous monitoring and human oversight to prevent failures and ensure accountability. Our framework suggests trust evolves with consistent interactions and AI reliability [57]. Initial deployments may start with limited autonomy, gradually increasing as the AI proves its competence. For example, a new malware detection model might begin with passive monitoring before taking an active role when accuracy is confirmed. Together with autonomy and HITL, trust completes the triad that enables safe, adaptive, and effective human–AI collaboration within modern SOCs.

### 4.4 Unified Representation of Autonomy, HITL, and Trust

The core elements essential for successful Human-AI collaboration within SOCs are *Autonomy, A*, *Human-in-the-Loop, H*, and *Trust, T*, and they are closely interrelated. Table 3 offers a comprehensive operational framework that demonstrates how varying levels of HITL align with SOC tasks, the complexity of tasks, AI autonomy, and their practical effects. This framework underscores the need for a dynamic equilibrium between human oversight and machine autonomy, emphasizing how trust and task complexity influence the necessary level of human engagement in AI-supported SOC activities. The interaction of these elements dictates the deployment of AI systems, the extent of control retained by human analysts, and the perceived reliability and usage of AI aid. Autonomy specifies how independently AI operates; HITL indicates the amount of human oversight; and trust influences both, facilitating the transition between human-led and AI-driven decision-making.

To operationalize this triadic relationship, we present a unified mapping across representative SOC tasks in Table 4. Each task is associated with typical ranges of autonomy ($A$), human involvement ($H = 1 - A$), and the level of trust ($T$) required for effective delegation.Table 4 highlights that high-autonomy tasks such as blocking known malicious IPs require low human intervention but high trust in the system's reliability. Conversely, novel or ambiguous tasks

Table 3. Operational Mapping of HITL with Task complexity and Autonomy in SOC

| HITL Level | SOC Task Example | Task Complexity | AI Autonomy | Operational Implications |
|---|---|---|---|---|
| Full HITL | Investigating zero-day threat | High | Low | Analyst-led decision-making |
| Partial HITL | Quarantine suspicious malware | Medium | Medium | Balanced control and oversight |
| HOtL | Auto-classify phishing emails | Low to Medium | High | AI-led with human supervision |
| HOoTL | Blocking malicious IPs, auto-patching | Low | High | Fully automated, routine tasks |

like investigating zero-day exploits require low autonomy and high human control, due to insufficient trust or high uncertainty. Tasks in between, like phishing classification, benefit from shared control, requiring a balance of moderate autonomy, oversight, and trust. By adjusting these three dimensions according to task characteristics, SOCs can enhance efficiency while ensuring accountability, interpretability, and critical human judgment. This coordination underpins the Human–AI Collaboration Framework

Table 4. Triadic Mapping of Autonomy, HITL, and Trust Across SOC Tasks.
**Legend:** $A$ = Autonomy (0 = manual, 1 = full autonomy), $H$ = Human-in-the-Loop (inverse of $A$), $T$ = Required trust level to enable safe AI delegation.

| SOC Task | Autonomy ($A$) | HITL ($H$) | Trust ($T$) |
|---|---|---|---|
| Blocking a known malicious IP | High (0.8–1.0) | Low (0–0.2) | High |
| Phishing email classification with Human review | Moderate (0.4–0.7) | Moderate (0.3–0.6) | Medium |
| Investigating zero-day exploit | Low (0.1–0.3) | High (0.7–0.9) | Low |

### 4.5 Situational Awareness- Human–AI Collaboration

Situational awareness (SA) is a foundational element of our proposed Human–AI Collaboration Framework for SOCs, enabling shared, context-sensitive decision-making across different levels of AI autonomy and analyst roles. Within HITL configurations, SA ensures that AI-generated insights are interpreted in line with human mental models, preserving accountability and relevance in operational decisions. As Aref et al. [5] emphasizes, this shared awareness between humans and AI enhances team cognition and improves detection accuracy, particularly in identifying stealthy or ambiguous threats. Following Endsley's model of perception, comprehension, and projection [14], SA in SOCs relies on Human-AI collaboration to detect, interpret, and predict threat trajectories. Aref et al. [5] stress that AI should not work in isolation, but rather communicate its reasoning and uncertainty clearly to aid analysts. Challenges include automation bias leading analysts to over-rely on AI, suppressing critical judgment [51], and cognitive overload due to excessive alerts, reducing vigilance [35]. These weaken situational awareness and the ability to identify high-priority threats. AI should adjust information delivery based on cognitive load, relevance, and urgency [5].

Building upon the A$^2$C framework by Baruwal Chhetri et al. [9], we conceptualize SA as an interactive feedback loop between AI capabilities, such as autonomous analytics and large-scale data processing, and human expertise, like domain knowledge and contextual understanding. This model breaks down SA into three interconnected components: *self-awareness, teammate awareness, and world awareness.* In this context, self-awareness involves both human and AI systems recognizing their limitations. Humans need to be aware of issues like fatigue or skill deficits, while AI must evaluate its confidence and yield when unsure. Teammate awareness is essential for fostering reciprocal understanding between agents, ensuring both are aware of each other's roles, cognitive states, and workload for effective collaboration.

World awareness entails keeping a current and coherent view of the operational environment, covering threat landscapes, system status, organizational objectives, and ongoing response measures.

In our framework (see Figure 3), situational awareness emerges as the central mediator of collaboration, dynamically shaped through continuous feedback between human and AI actions. It enables seamless transition across HIL configurations, aligned with varying levels of autonomy and SOC tiers. This alignment facilitates role-adaptive information exchange, enhances analyst trust, and supports timely decision-making in high-pressure environments.

### 4.6 SOC Role Tiers: Functional Responsibilities and AI Alignment

This section outlines SOC roles aligned with AI within our framework. Security Operations Centers use a multi-tier approach for assigning analyst duties by complexity, expertise, and risk. Tiers 1 through 4 align with the NIST CSF functions, forming the base of our Human–AI Collaboration Framework. This setup integrates AI autonomy with human oversight for effective cyber defense. In a ransomware attack starting with an endpoint breach, SOC responses are managed across these tiers, using varying levels of AI and human collaboration.

Tier 1: System Hardening and Alert Monitoring (Identify/Protect). Attacks are initially detected by Tier-1 analysts monitoring security telemetry and endpoint behavior. At low AI autonomy (Levels 0–1), analysts manually inspect alerts using SIEM and EDR systems. As autonomy increases (Level 2), AI tools assist in event correlation, anomaly detection, and alert prioritization. In HITL mode, analysts validate pre-processed alerts and initiate playbooks, ensuring quick triage without compromising judgment.

*Tier 2: Threat Investigation and Coordinated Mitigation (Detect/Respond).* Upon escalation, Tier-2 analysts investigate the threat and initiate containment procedures such as isolating compromised systems. At autonomy Levels 2–3, AI enriches investigations with contextual data, asset relationships, and suggested mitigation steps. Analysts supervise SOAR workflows in HOtL mode, adapting automated actions to operational context. This supports timely containment while preserving control and oversight.

*Tier 3: Proactive Threat Hunting and System Optimization (Detect/Recover).* Tier-3 analysts explore the attack's scope, identifying attacker lateral movement and persistent footholds. AI systems at higher autonomy (Levels 3–4) proactively detect similar behavior across the network and generate threat hypotheses. Analysts validate insights, tune detection logic, and manage AI feedback loops, improving future detection and supporting organizational recovery.

*Tier 4: Governance, Oversight, and Strategic Risk Management (Govern/Recover).* Tier-4 personnel assess policy and regulatory implications, coordinate incident reporting, and update organizational response frameworks. At this level, AI supports tasks such as audit logging, compliance validation, and policy review. Operating in HOoTL mode, analysts validate AI-generated reports and enforce standards on explainability, risk accountability, and legal compliance.

This mapping demonstrates how our framework facilitates smooth collaboration between humans and AI within SOC tiers. By matching functional duties with autonomy levels and HITL setups, our framework aims to improve Mean Time to Respond (MTTR), boost situational awareness, and increase trust in AI-supported tasks, all while ensuring transparent human accountability and supervision.

### 4.7 Collaborative SOC Workflows

The proposed Human–AI Collaboration Framework supports dynamic, tiered workflows across SOC functions, adapting to varying levels of AI autonomy and trust. As AI capabilities scale, task delegation shifts from manual analyst-driven processes to conditionally or fully autonomous execution, with human analysts maintaining oversight and strategic control. Figure 4 depicts collaborative SOC workflows grounded in our proposed framework, aligned with three primary

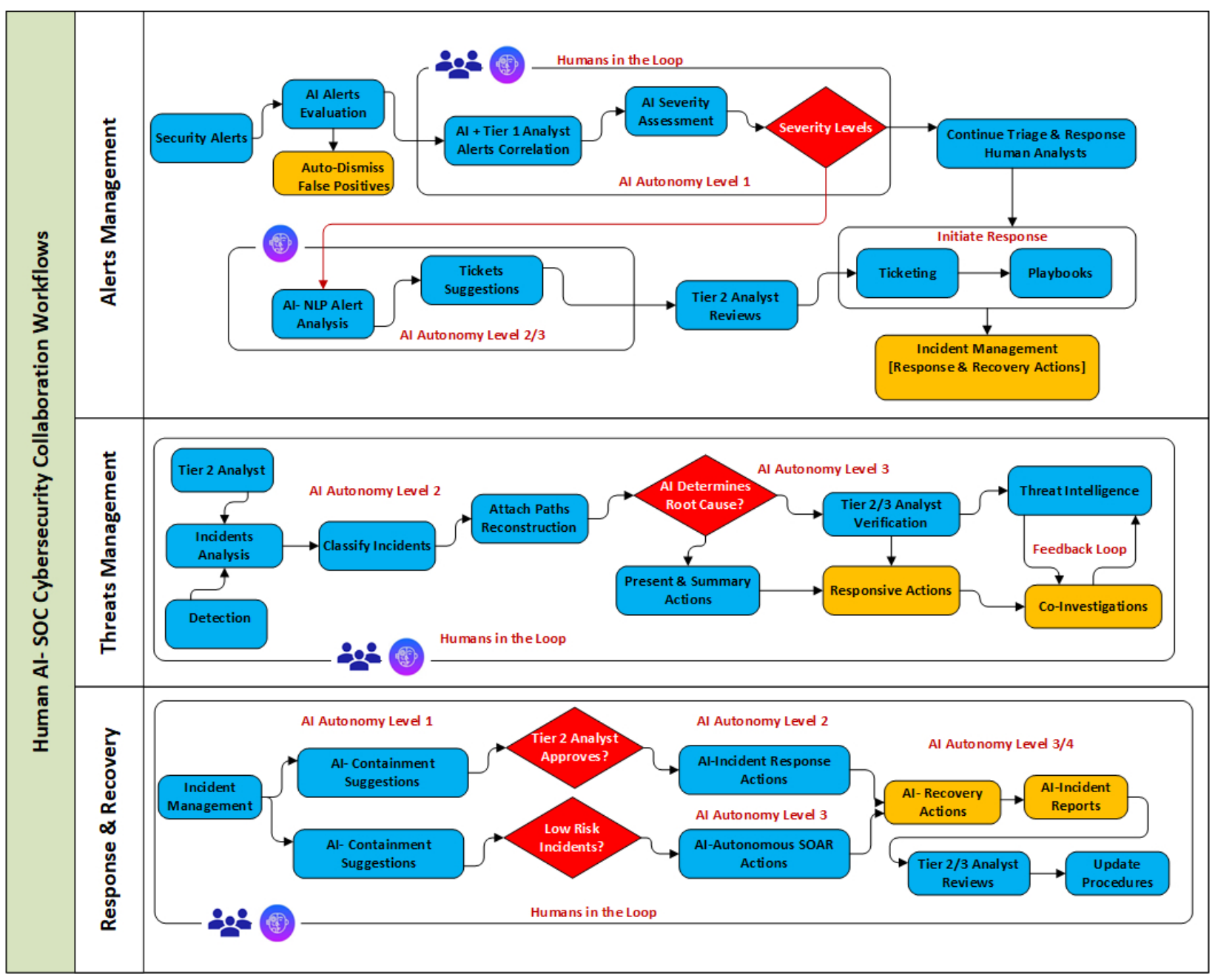

Fig. 4. Operational Workflows of Human–AI Collaboration in SOCs: AI autonomy levels guide how alerts, incidents, and response actions are shared and executed between analysts and AI agents across key functional domains.

SOC functions: *(i)* Alerts Management, *(ii)* Threats Management, and *(iii)* Response & Recovery. For simplicity, we have excluded system protection, as its processes are fairly straightforward and align well with the designated roles for humans and AI. Each domain illustrates the role of AI in assisting or automating essential steps based on the degree of autonomy and task complexity.

*(i) Alerts Management* The Alerts Management workflow initiates with AI agents operating at Level 1 autonomy, assisting Tier-1 analysts by correlating alerts, performing preliminary severity assessments, and supporting triage decisions. Humans remain actively involved in validating the AI's output and determining the appropriate response path based on alert severity. At Levels 2 and 3, AI systems advance to auto-dismiss false positives using contextual patterns, aided by NLP-based alert analysis. These systems generate ticket suggestions and recommend response actions, which are reviewed by Tier-2 analysts. At this stage, AI not only clusters and summarizes related alerts into incident candidates but also refines its prioritization through continuous feedback loops. Despite increased autonomy, human analysts remain "in-the-loop" to automatically dismissed alerts, validate critical escalations, and ensure alignment with operational context. Human judgment is especially vital in ambiguous cases and for triage escalation. This layered collaboration improves detection fidelity, reduces analyst fatigue, and fosters trust through transparency, iterative validation, and autonomy scaling aligned with risk.

*(ii) Threats Management* In the Threats Management workflow, AI agents operate at Level 2 autonomy to assist Tier-2 analysts in classifying incidents and reconstructing attack paths through automated log correlation and contextual analysis. This reduces investigation time and supports evidence collection for emerging threats. At Level 3, AI performs root cause analysis and generates summary reports with proposed responsive actions. Tier-2 and Tier-3 analysts remain "in-the-loop," verifying AI conclusions, incorporating operational context, and resolving ambiguities, such as differentiating legitimate administrative activity from malicious behavior. Verified intelligence is then used to drive mitigations and update detection models. This co-investigative loop driven by human validation and AI-generated insights, ,enhances investigative depth, accelerates incident resolution, and supports adaptive learning. Feedback mechanisms between analysts and AI systems ensure continuous model refinement and increased precision in threat characterization and response planning.
*(iii) Response & Recovery* The framework adopts a tiered autonomy approach to streamline containment, response, and recovery workflows. At Level 1 autonomy, AI recommends containment actions (e.g., host isolation, account lockdown), which require explicit approval from Tier-2 analysts. As trust builds, Level 2 enables AI to autonomously execute predefined incident response actions via SOAR integration, following analyst confirmation for moderate-risk events. At Level 3, AI agents autonomously run SOAR playbooks for low-risk incidents, escalating only ambiguous or critical cases for human review. Analysts retain oversight and can intervene through real-time monitoring and override controls. This configuration significantly reduces response time and mitigates damage from fast-moving threats. At Level 4, AI agents manage end-to-end recovery operations, including post-incident system restoration and automated reporting. Human analysts shift to supervisory roles, overseeing policy alignment, compliance, and procedural updates. In parallel, advanced AI-driven threat hunting capabilities at Level 3 use machine learning, anomaly detection, and graph-based analysis to proactively uncover latent threats. Outputs are validated by Tier-3 analysts, with feedback, such as false positive tags, used to iteratively fine-tune AI models. This continuous collaboration enables rapid, scalable, and trustworthy recovery and threat management workflows.

These workflows showcase Human–AI collaboration, balancing autonomy and oversight to improve SOC performance and reduce analyst workload. Alongside SOC tasks, workflows like ticketing, change management, and integrated risk management should be included to align cybersecurity with enterprise goals, strengthening accountability, resilience, and coordination across functions.

Table 5. Comparative analysis of related work on AI in SOCs and Human-AI collaboration. Legend: ✓= partially supports; ✓✓= moderately/full supports; ✓✓✓= comprehensively supports; ✗= not addressed. SOC Functions: Sys. Mon. = System Monitoring (Protection + Ingestion & Log Mgmt); Thr.Mgt = Threat Management (detection/hunting); Alrt Mgt. = Alerts Management (triage/analysis); IR&R = Incident Response & Recovery. Autonomy: primary autonomy level or range targeted (0 = manual … 4 = full AI).

| Autonomy | Sys. Mon. | Thr.Mgt. | Alrt Mgt. | IR&R | Human-AI Collaboration | Authors (Year) |
|---|---|---|---|---|---|---|
| | SOC Functions | | | | | |
| Level 0–3 | ✗ | ✓ | ✓ | ✓ | Human factors, LOA framework | Tilbury & Flowerday (2024) |
| Level 1–3 | ✗ | ✓ | ✓✓ | ✓ | Autonomy, collaborate SA | Chhetri et al. (2024) |
| Level 1–2 | ✓ | ✓✓ | ✗ | ✗ | Reciprocal Learning – HITL | Cohen et al. (2025) |
| Level 3 | ✗ | ✓✓ | ✗ | ✓ | HITL, cognitive modelling | Aref et al. (2025) |
| Level 0–4 | ✓✓✓ | ✓✓✓ | ✓✓✓ | ✓✓✓ | Autonomy, Trust, HITL, and SA | **Our work** |

## 5 Related Work and Comparative Analysis

There is a growing body of research on introducing AI into SOC operations and on frameworks for human-automation collaboration. Here we review and compare related works from the literature, focusing on how they address core SOC functions, levels of autonomy, and Human-AI teaming or collaboration aspects. Table 5 summarizes key works in terms of their coverage of SOC security functions (Protection, Ingestion & Logging, Threat Detection/Management,

Alerts Management, Incident Response/Recovery), the autonomy level they primarily target, and their consideration of Human-AI collaboration (including any frameworks or models proposed).

Prior studies highlight AI's role in enhancing analyst efficiency with human oversight in SIEM and SOAR, facilitating semi-automated decision-making [21, 23, 47]. Jalalvand *et al.* [19] studied AI-driven alert prioritization in SOCs through adaptive, explainable models using contextual and historical data. Chhetri *et al.* in [9] proposed a Human-AI framework emphasizing shared awareness and trust to reduce alert fatigue and improve decision efficiency. Tilbury *et al.* [51] developed an automation approach for cybersecurity, introducing a SOC automation matrix for balanced Human-AI collaboration. Most SOCs are minimally automated, heavily dependent on analysts, with recommendations against full automation, supporting gradual transitions to human-in-the-loop operations. Binbeshr *et al.* [7] highlighted AI's prevalence in SOCs, as machine learning offers better accuracy and fewer false positives, but faced challenges in data quality and model interpretability, necessitating human oversight. Tilbury and Flowerday [50, 51] perform a scoping review on *human-automation augmentation in SOCs*, identifying automation areas and issues like automation bias and analyst complacency. Their key contribution, the SOC Automation Matrix, categorizes four human-automation relationship levels and incorporates Levels of Automation (LOA). They emphasize that while automation can cover the incident lifecycle, managing socio-technical factors (like analyst trust) is essential, advocating for human-in-the-loop controls and offering guidance on aligning automation levels with tasks. Our framework complements their work by centralizing trust in Human-AI collaboration.

On the frameworks side, Chhetri *et al.* [9] present the $A^2C$ (Automate-Augment-Collaborate) framework to tackle alert fatigue. It enables dynamic switching between three modes: automated (AI handles routine alerts), augmented (AI offers suggestions to speed up analyst decision-making), and collaborative (human and AI jointly explore complex threats). Their vision is a flexible SOC where the level of AI autonomy can dial up or down per context, which resonates with our approach. They report that implementing $A^2C$ in a prototype reduced false-negative incidents while keeping analysts in control for novel situations. Another notable work is by Cohen *et al.* [11], who focus on *Human-AI collaboration for cyber threat intelligence (CTI).* They introduce a Reciprocal Human-Machine Learning (RHML) model wherein two analysts and an AI system iteratively learn from each other to classify hacker forum messages. Over time, both detection accuracy and the human experts' understanding of the threat landscape improved. This demonstrates the power of continuous learning in Human-AI teams and provides a concrete example of human-in-the-loop machine learning in SOC-like tasks. Aref et al. [5] developed a Cognitive Hierarchy Theory-driven Deep Reinforcement Learning (CHT-DQN) framework for modeling Human-AI collaboration in cloud SOCs. It integrates cognitive modeling, attack graphs, and HITL design for real-time threat mitigation. However, the focus remains primarily on threat management, with limited attention to alert management, system monitoring, and incident response.

Our framework builds on existing research by presenting a model that combines various autonomy levels (0-4), human roles (HITL), and collaboration principles for key SOC functions, focusing on trust balance and shared situational awareness. It covers a wider range of SOC tasks, from alert management to incident response and recovery, unlike many studies that focus on specific elements like alert management or scenarios such as CTI and vulnerability management. Our approach integrates these aspects into a unified Human-AI collaboration for SOCs. We demonstrate its practical value through exemplification of an empirical case study (Section 6).

## 6 Case Study: ACDC CyberAlly for SOC

To demonstrate the practical application of the proposed Human–AI Collaboration Framework in a controlled yet realistic environment, we conducted a case study utilizing a cyber range-based SOC simulation developed under the

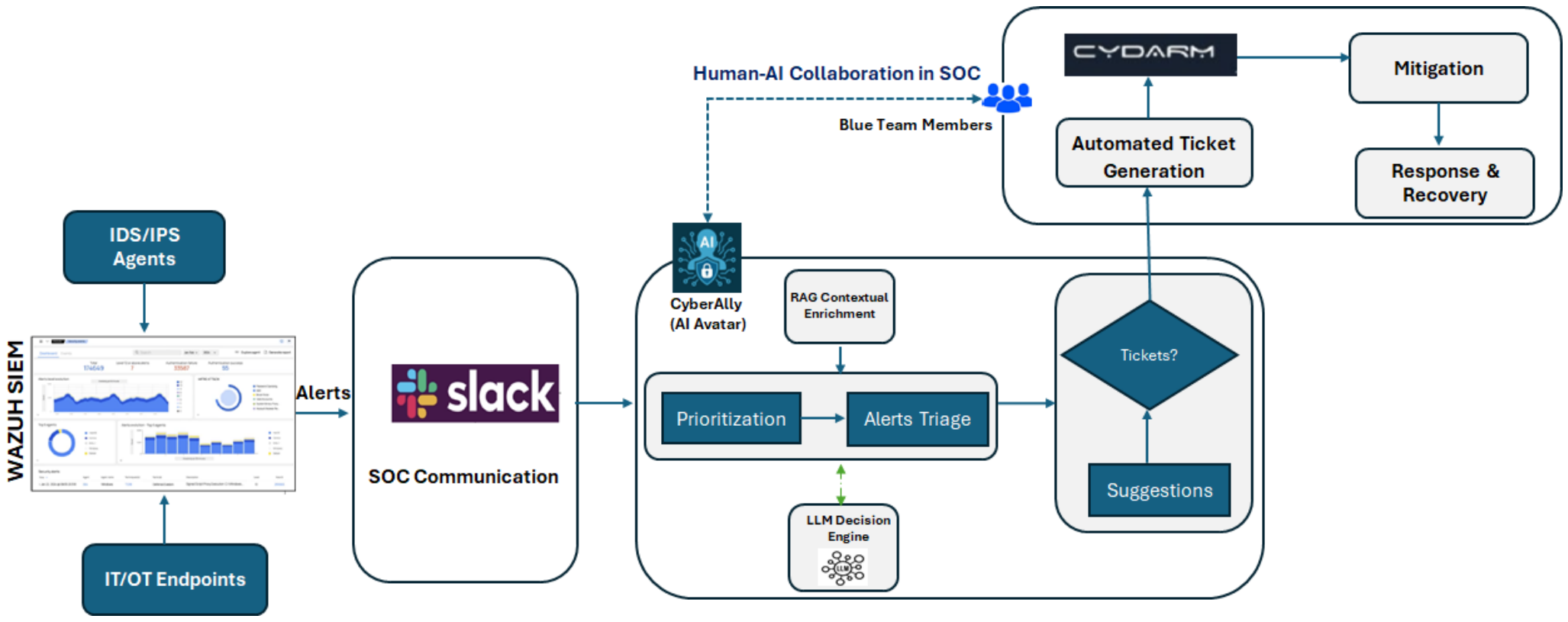

Fig. 5. Human–AI collaboration in a simulated SOC environment integrated into the ACDC Cyber Range. The CyberAlly (Avatar) AI-augmented workflow supports the Blue Team by combining real-time alert triage,LLM-based analysis, and automated ticketing.

*Augmenting Cyber Defence Capability (ACDC)* project. The system, termed **CyberAlly**, functioned as an AI-augmented security analyst embedded within a red team (attacker) versus blue team (defender) exercise. This section outlines the simulation design, attack scenarios, human–AI collaboration dynamics, SOC performance outcomes, and key insights derived.

### 6.1 Human-AI collaboration: Wargames Simulation Setup

A representative SOC simulated environment was designed, replicating the operational setup of a mid-sized enterprise network. The environment integrates core SOC components, including a SIEM aggregating security telemetry from servers, endpoints, and network devices; host and network-based intrusion detection systems; firewalls; an endpoint detection and response (EDR) suite; and a SOAR platform enabling automated incident response workflows. On top of this infrastructure, we deployed CyberAlly, an AI avatar agent built upon a fine-tuned large language model (LLM) architecture (based on OpenAI's GPT models), customized for cybersecurity operations. CyberAlly was engineered to interface natively with SOC tools: querying the SIEM, initiating SOAR workflows, and communicating with analysts via a dedicated chat console. CyberAlly's capabilities were enhanced with domain-specific training using two years of simulated cyber incidents and data from five ACDC cyber defense exercises [22], each structured as wargames involving 4 Red Team and 10 Blue Team members simulating adversary-defender dynamics. A *Retrieval-Augmented Generation (RAG)* mechanism allows dynamic queries to a security knowledge base with MITRE ATT&CK tactics, threat intelligence feeds, and network topology. Knowledge graphs map critical entities like hosts, users, vulnerabilities, and incidents, minimizing LLM errors and ensuring verified outputs. Blue Team Analysts use a Slack SOC dashboard for interactions, receiving insights, and anomaly notifications like unusual logins. As a SOC tier analyst, CyberAlly manages events and collaborates with humans at Level 2 autonomy, requiring approval for major actions.

Figure 5 depicts the workflow of the CyberAlly-augmented SOC. Security alerts from IT/OT endpoints and IDS/IPS agents are ingested via Wazuh and routed through a central communication hub (Slack), ensuring visibility for human analysts and the CyberAlly AI assistant. CyberAlly performs automated prioritization and triage, leveraging contextual enrichment from knowledge graphs (via RAG) and LLM-driven analysis to generate actionable insights and recommendations. Alerts classified as benign are archived, while suspicious or critical incidents proceed to collaborative review; analysts can approve, escalate, or provide feedback, continuously refining the AI's performance. CyberAlly facilitates automated ticket creation in the case management system (Cydarm) for high-severity cases, streamlining mitigation and recovery actions. All decisions, tickets, and analyst feedback are systematically logged and fed back into the knowledge base, closing the loop for adaptive learning and operational improvement across both AI modules and human workflows.

**Attack Scenarios**

During the ACDC cyber wargames, three attack scenarios disrupted maritime port IT and OT systems. The Red Team used the cyber kill chain and MITRE ATT&CK frameworks in the ACDC SOC, with AI support from the CyberAlly AI Avatar. The hybrid IT-OT ACDC Port system includes 55 VMs for business, public services, IoT devices, and ICS logistics. Here are brief descriptions of the simulated scenarios:

*Scenario 1: IoT Depth Sensors Attack.* The first scenario involves an attack on the port's 20 km channel operations, reliant on IoT-based depth sensors for vessel coordination. These sensors collect tidal data and share it with stakeholders, the harbor master, and shipping companies via a web portal. The Red Team aims to compromise IoT depth sensors and backend systems to inject false data and disrupt transmissions, risking unsafe navigation or halting operations. The Blue Team ensures IoT data integrity and availability and maintains services with the Weather Central server.

*Scenario 2: Conveyor Belt Attack.* The second scenario involves compromising the port's OT loading system, specifically the conveyor belt used for loading iron ore. Controlled by OT components (PLCs, HMIs, and SCADA)[2] and integrated with IT networks, the Red Team attempts to manipulate system logic to disrupt OT controls, causing safety hazards and loading delays. This underscores the protection challenges of mission-critical OT assets interconnected with enterprise services.

*Scenario 3: Autonomous Train Supply Chain Attack.* The final scenario targets the autonomous train system transporting materials to the port. Managed by an OT system controlling speed, braking, and unloading, the Red Team seeks to derail train operations, interrupt unloading, and affect supply chains. This scenario underscores cybersecurity concerns in interconnected, automated port operations.

### 6.2 Performance Analysis of Human–AI Collaboration in SOC Tasks

***Detection and Triage with CyberAlly.*** Operating as an LLM-enhanced AI agent within the simulated ACDC Cyber Range SOC, CyberAlly continuously monitored a high volume of incoming alerts from ***SIEM (Wazuh)*** and EDR systems. Leveraging its LLM capabilities, CyberAlly not only ingested and filtered events in real-time but also performed advanced classification, prioritization, and severity scoring, which enabled faster and more accurate detection of incidents. This intelligent triage workflow provided foundational support for automated ticket generation and SOAR playbook activation, drastically improving response times and analyst efficiency. Initially, CyberAlly served as an intelligent alert filter and advisor at Level 1 *(Full HITL)*, monitoring data streams and flagging anomalies. In one attack scenario, a phishing email targeting IT enterprise users was used by the Red Team to establish a reverse shell, which then facilitated lateral movement into the OT engineering workstation. CyberAlly auto-classified the phishing emails as malicious, prompting human analysts to verify its decisions while trust was being established. Despite this initial caution, CyberAlly significantly reduced false-positive alerts, enabling analysts to focus on high-fidelity threats. Its context-aware alert prioritization, such as recognizing anomalous login attempts for OT admin accounts from foreign

[2]These are Maritime Port system OT assets part of wargame exercises.

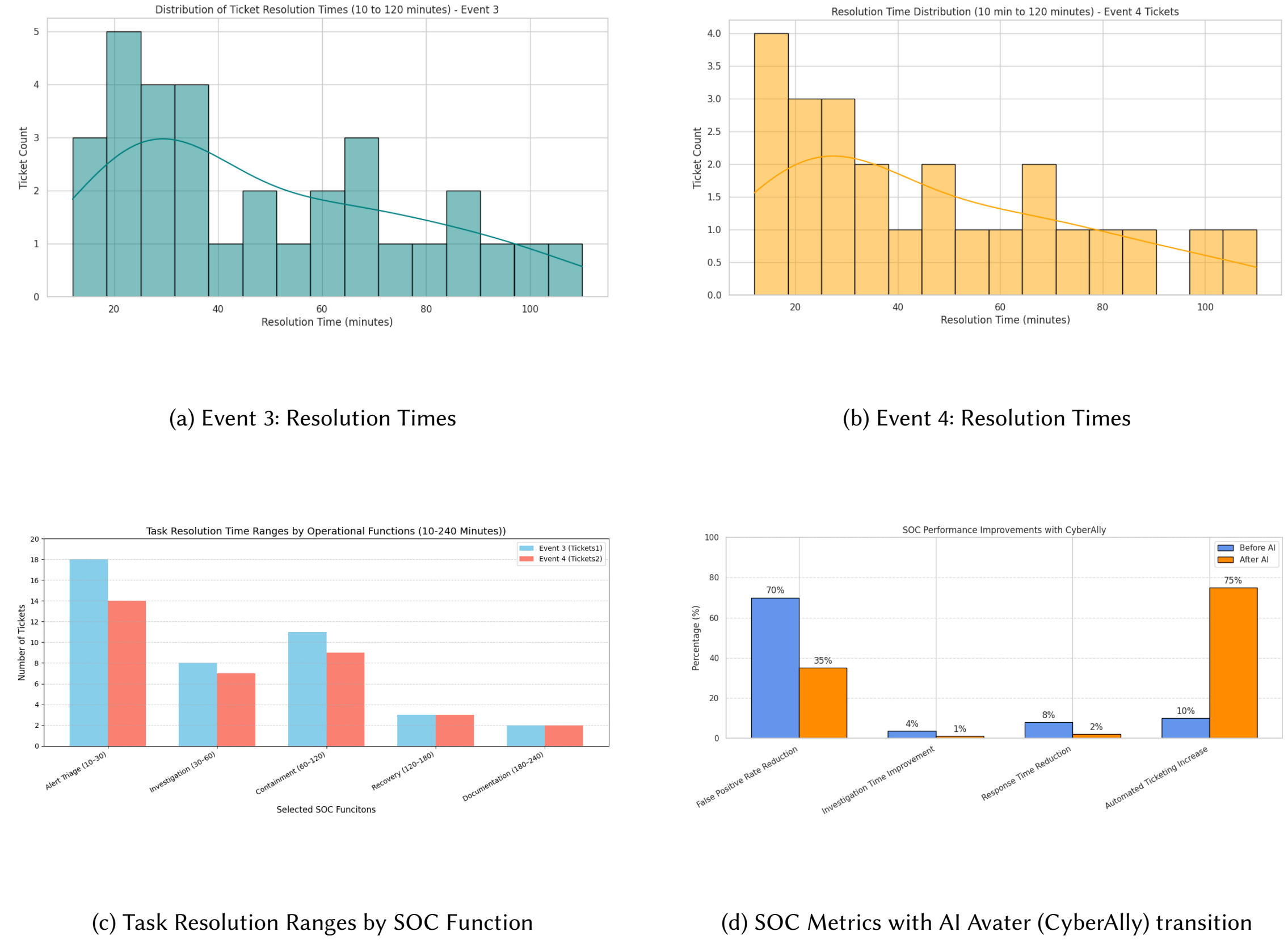

(a) Event 3: Resolution Times

(b) Event 4: Resolution Times

(c) Task Resolution Ranges by SOC Function

(d) SOC Metrics with AI Avater (CyberAlly) transition

Fig. 6. Visualizing SOC performance with CyberAlly across response times, task types, and operational gains during wargame events.

IPs at 3 PM, exemplified the Human-on-the-Loop paradigm—routine alerts were autonomously filtered, while critical anomalies were escalated.

Figures 6a and 6b show ticket resolution distributions for Events 3 and 4, respectively. In Event 3, most incidents were triaged and mitigated within 20–40 minutes, underscoring CyberAlly's utility in rapid alert classification and early-stage containment at Tier-1 and Tier-2 levels (Level 2 autonomy). In Event 4, despite more sophisticated threats such as a false data injection attack on IoT depth sensors, where attackers altered tide measurement values to disrupt port channel operations, resolution time again clustered below 40 minutes. This highlights CyberAlly's effectiveness in anomaly detection and recommending validated mitigation responses. While analysts maintained control over decision-making, AI-driven triage halved false positives from 70% to 35%, significantly improving the signal-to-noise ratio.

***Investigation and Contextual Correlation.*** In cases like credential misuse, unauthorized OT access, and PLC and HMI manipulation during conveyor belt operations, CyberAlly operated semi-autonomously ***(Level 2)***. It automatically retrieved logs, checked historical incidents, and mapped Red Team actions to MITRE ATT&CK tactics. In the conveyor belt MITM scenario, CyberAlly traced the attack from an IT phishing breach to an OT network compromise via known techniques like credential dumping and remote command execution. It proposed actions like checking OT workstation logs or verifying PLC setups. While ***Tier-2 analysts*** retained authority over final decisions, CyberAlly automated repetitive and cognitively intensive tasks like log correlation and hypothesis generation. This automation cut investigation time from three hours to one (a 67% improvement), as shown in Figure 6d. The AI's ability to leverage both RAG modules and dynamic knowledge graphs enriched the investigation process with asset criticality insights, such as highlighting that compromised conveyor belt PLCs directly impacted vessel loading operations. Analysts reported that CyberAlly often spotted subtle associations they missed, such as a web server access pattern linked to a known CVE or lateral movement hinting at a planned OT compromise. While autonomous actions like service restarts still required approval, the AI's contextual guidance significantly accelerated root cause analysis and action planning.

Figure 6c reveals that the majority of incidents in both events were resolved during triage (10–30 minutes) and containment (30–120 minutes). CyberAlly's integration of RAG and internal knowledge graphs enabled faster incident analysis. Human analysts, especially at Tier-2, benefited from CyberAlly surfacing relevant MITRE ATT&CK tactics and incident analogs, streamlining investigative workflows across diverse threats, including train supply chain disruptions and OT system compromises.

***Response and Recovery Efficiency Gains.*** With analysts' trust, CyberAlly was allowed more *autonomous* actions under *human-on-the-loop conditions*. In the train supply chain scenario, it detected abnormal commands affecting OT components for speed regulation and unloading. CyberAlly promptly alerted engineers, flagged signals, and isolated PLCs before suggesting a system-wide inspection. In loading and web-server attacks, it used SOAR integration to rapidly isolate endpoints and block malicious IPs, reducing MTTR from 8 hours to 90 minutes. Analysts supervised and adjusted AI decisions, such as reversing unintended blocks. Figure 6d shows CyberAlly's impact, increasing automated ticketing from 10% to 75% and cutting response times by over 60%. In complex incidents like train unloading or conveyor manipulation, it followed predefined steps under HITL, boosting SOC agility and resilience.

***Autonomy Calibration and Trust Evolution.*** From a Human-AI collaboration perspective, the integration of AI Avatar (cyberally) into simulated SOC operations was aligned with various HITL configurations based on incident complexity. The early stages of triage employed full HITL, ensuring each AI-flagged alert was human-reviewed. As CyberAlly demonstrated accuracy, particularly in classifying phishing emails and triaging anomalies in OT systems, its role evolved to *on-the-loop supervision* for routine classifications. In later response phases, especially when isolating endpoints or quarantining known threats, the system operated with partial or even full autonomy under specific rules of engagement. This calibrated autonomy approach ensured efficient task delegation without sacrificing human oversight. Trust in CyberAlly was not assumed; it was earned. Analysts initially scrutinized all AI outputs. Over time, explainable decisions, high-accuracy classifications, and consistent performance encouraged analysts to treat the AI as a reliable collaborator. By the final scenario, CyberAlly autonomously managed most routine SOC workflows, allowing human analysts to focus on complex decision-making. Misclassifications were rare and leveraged as learning points, further refining AI outputs.

### 6.3 Lessons Learned

The lessons learned from the ACDC CyberAlly simulation validate key principles from our Human–AI Collaboration Framework. They highlight how trust calibration, dynamic autonomy scaling, and structured human-in-the-loop configurations collectively enhanced SOC performance. These insights offer practical guidance for deploying AI-assisted security operations in real-world environments.

(1) *Autonomy Calibration Across SOC Functions.* A gradual, tiered escalation of CyberAlly's autonomy proved essential for balancing operational efficiency with human oversight. Early triage phases employed full HITL validation, while later response actions transitioned toward partial or conditional autonomy. Routine tasks such as phishing triage and endpoint isolation were successfully automated, while critical incident responses remained under human supervision. This autonomy layering fostered analyst trust, optimized AI deployment, and minimized risks of automation bias or decision errors, highlighting the importance of dynamic autonomy calibration based on incident complexity and trust levels.

(2) *Human–AI Communication and Autonomy.* Initially, analysts struggled with CyberAlly's queries and responses but adapted by refining their language and strategies. This highlights the need for Human-AI co-adaptation in SOC workflows. As AI autonomy increased, communication evolved. CyberAlly started at Level 1 autonomy, needing human verification, and progressed to Level 2 for phishing and alerts, reaching conditional Level 3 for low-risk tasks under supervision. These shifts matched SOC task complexity, proving that autonomy scaling should be adaptive, trust-based, and task-specific, ensuring efficiency while keeping human oversight in critical decisions.

(3) *Trust but Verify.* Analysts in the early stages approached CyberAlly's AI-suggested alerts with caution, manually verifying everything, aligning with Level 1 autonomy (Full HITL). Over time, as CyberAlly proved accurate in alert triage and threat prioritization, trust grew, leading to Level 2 autonomy, where analysts only intervened when necessary. CyberAlly's transparent justifications were crucial for building trust. This progression showed that SOC autonomy should adapt to AI reliability.

(4) *Avoiding Information Overload.* Rather than overwhelming analysts, CyberAlly reduced cognitive load by filtering out benign noise. However, this positive outcome required careful calibration of alerting thresholds and relevance criteria. Without tuning, an AI assistant could exacerbate alert fatigue. Iterative adjustment cycles are necessary to optimize signal-to-noise ratios based on operational contexts.

(5) *Effective Training is Essential.* Extensive domain-specific fine-tuning of CyberAlly using cyber range logs, security playbooks, and knowledge graphs was crucial. While the base LLM model was fluent, it made inaccurate assumptions about security events until retrained with operationally relevant data. This experience reinforces that AI assistants in SOCs must be tailored to the specific organizational context; relying on general-purpose models risks unreliable performance. Organizations adopting AI SOC assistants must prioritize secure, privacy-compliant enrichment of the AI with their internal datasets.

The ACDC CyberAlly case study showed the effectiveness of Human-AI Collaboration in SOCs. Combining AI's speed with human reasoning, the team reduced alert fatigue, sped up responses, and improved decision-making. The AI agent (CyberAlly) handled routine tasks, allowing analysts to focus on complex threats. Human oversight was crucial for interpreting ambiguities and maintaining ethics. The study emphasized trust calibration, explainable AI, training, onboarding, and incident learning. AI supports, not replaces, human expertise, addressing analyst overload and aligning with cybersecurity standards. The CyberAlly simulation demonstrated Human-AI collaboration for resilient SOCs.

## 7 Discussion

### 7.1 Framework Contributions

Central to the framework is a taxonomy of autonomy levels, defining the division of tasks between AI and humans in SOC operations. The framework promotes dynamic task allocation: routine, high-volume activities are delegated to AI agents, while humans retain control over complex decision-making and strategic judgments. As incidents become more sophisticated or the analyst workload increases, the framework supports real-time adjustment of roles, progressively shifting autonomy to AI under supervised conditions. Task handovers are carefully managed: AI initially operates under strict HITL supervision (Level 1 autonomy) and can be gradually elevated to more autonomous roles (Level 2- 4) as reliability and trust grow. Trust management is crucial to the framework, and trust is established through HITL and autonomy levels gradually. At low autonomy levels, human analysts verify each AI recommendation, shifting at higher levels to focus on critical actions. The framework requires AI to provide clear, evidence-backed explanations, enabling rapid assessment of AI outputs. This includes transparency about data sources, threat indicators, and uncertainty estimates, ensuring safe scalability of autonomy with human oversight. The modular framework integrates into diverse SOC environments without full system replacements, fitting within budgetary and legacy constraints. It blends cognitive systems engineering, adaptive autonomy, and cybersecurity into a practical SOC architecture, formalizing autonomy levels for cybersecurity and addressing gaps like trust calibration, human–AI co-learning, and modular deployment.

A case study in the ACDC cyber range demonstrated our framework by simulating a SOC with red/blue teams. We incorporated a refined LLM agent, CyberAlly, into the SOC's security stack, interfacing with SIEM (Wazuh), SOAR, and a knowledge base grounded in MITRE ATT&CK. Operating at Level 2 autonomy, CyberAlly monitored events, enriched context, and communicated with analysts through natural language. While analysts retained control over key actions, CyberAlly managed low-risk tasks like triage and event enrichment. This led to fewer false positives **50%**, faster investigations **67%**, and reduced MTTR **80%**. Analysts had less cognitive load, concentrating on strategic tasks like threat hunting. Feedback lauded CyberAlly's understandable outputs for building trust. The study affirms that increased autonomy, transparency, and oversight boost SOC efficiency, benefiting other SOCs with AI copilots.

### 7.2 Broader Implications for SOC Cybersecurity Operations

The adoption of structured Human–AI collaboration frameworks such as ours carries broader implications for cybersecurity operations. Operationally, it provides a pathway to manage the scale and complexity of modern cyber threats, where human only SOCs are increasingly overwhelmed. By automating repetitive tasks and augmenting situational awareness, AI enables faster threat detection, quicker containment, and more effective recovery, ultimately reducing breach impacts and downtime. Strategically, human–AI teaming addresses systemic issues such as the cybersecurity workforce shortage. By augmenting analyst capabilities and reducing burnout, AI-assisted SOCs can retain skilled personnel longer and enable less-experienced analysts to perform at a higher level. Furthermore, the continuous feedback mechanisms embedded in the framework promote the development of organizational knowledge over time, allowing SOCs to evolve and adapt more quickly to emerging threats. The trust calibration mechanisms proposed particularly the requirement for evidentiary AI outputs—help manage the sociotechnical risks of AI adoption, such as automation bias, distrust, and overreliance. Maintaining human-in-command structures ensures that critical decisions, such as threat escalation, service takedown, and enterprise-wide containment actions, remain under human oversight, preserving accountability and ethical governance. Importantly, our framework positions the SOC not as a reactive entity but as a learning, adaptive system. By embedding co-learning mechanisms between humans and AI, the SOC becomes capable of evolving its defenses dynamically, preparing not just for known threats but for novel attack patterns that may emerge.

### 7.3 Limitations

The framework offers solid foundations for integrating human expertise, AI, and cybersecurity in modern SOCs. By focusing on autonomy, trust, and human-in-the-loop decision making, it suggests a scalable model for future cognitive SOCs. Nevertheless, challenges like security operation complexity, evolving AI systems, and human cognition must be addressed. These issues are critical for real-world deployment, where technical, organizational, and operational factors can affect Human–AI collaboration effectiveness in SOC workflows.

Continuous *model retraining*, required to incorporate evolving threat intelligence, raises governance questions about responsibility and operational reliability. Managing *model drift*, where system performance may degrade over time, also necessitates robust analyst feedback mechanisms to sustain contextual relevance. The complex nature of SOC systems, characterized by *heterogeneous* security tooling, varying telemetry standards, and fragmented data sources, adds further difficulty. Moreover, disparities between human and AI *cognitive processes*, including intuition, contextual judgment, and explainability, make seamless coordination non-trivial. These structural and *cognitive asymmetries* create challenges in aligning AI behavior with analyst expectations and operational intent. Domain-specific adaptation also presents a challenge. While the case study example demonstrates the framework's efficacy in a simulated SOC environment, real-world deployments must account for infrastructure diversity, workflow variation, and data quality disparities. Tuning models to each SOC's operational and contextual nuances is essential for performance generalization. Techniques like continual learning or federated adaptation can support this, although they require mature data engineering practices.

In terms of operational integration, deploying the Human-AI collaborative systems in active SOC environments introduces *engineering and interoperability* challenges. AI systems must be embedded within a diverse and often legacy tool ecosystem that includes SIEM, SOAR, and EDR platforms. Secure, real-time data exchange and explainable interfaces add complexity. Resource-constrained organizations may struggle with integration and maintenance, affecting scalability despite modularity. Trust and interpretability are crucial for effective Human-AI collaboration. The framework emphasizes *explainable AI*; translating model outputs into actionable, cognitively usable insights is still an open research question. The risk of either over-reliance or underutilization of AI outputs persists if trust is not calibrated through *transparent reasoning* and uncertainty estimation. Analyst-facing tools must evolve to support rapid validation of AI decisions, especially under cognitive load.

From an organizational perspective, human-centric adoption barriers, such as *role displacement* concerns or unfamiliarity with AI capabilities, can slow uptake and smooth integration in the current SOC ecosystem. Building trust over time through iterative deployment, training, and shared decision-making is key. Moreover, the effectiveness of the framework depends on sociotechnical readiness, including workforce AI literacy and cross-functional collaboration between analysts, engineers, and governance teams.

## 8 Conclusion and Future Work

This paper presented a structured framework for Human–AI collaboration in SOCs, addressing limitations in current automation-centric approaches. The framework facilitates scalable decision-making by coordinating AI autonomy, HITL, and Trust with analyst roles within SOC tiers, and adjusting trust via explainable and adaptive methodologies, ensuring human oversight is maintained. A case study highlighted concrete advantages, including diminished alert fatigue and enhanced response coordination. This research provides a solid foundation for developing next-generation Cognitive SOCs, where AI acts not as a substitute but as a dependable enhancement to human expertise in intricate cyber defense tasks.

Future research should enhance the Human–AI SOC framework by expanding domains that improve collaboration in cybersecurity. Next-generation SOCs may leverage AI agent teams for tasks like intrusion analysis and threat hunting, coordinated by an orchestrator agent [10, 25]. Research should address agent communication, redundancy reduction, and modeling of multi-agent workflows. While industry applications show potential, rigorous simulations and frameworks are needed to assess distributed AI in SOC contexts. Large Language Models (LLMs) are promising for tasks such as log summarization and incident reporting. Future work should emphasize domain-specific fine-tuning to improve safety and performance. Tools like cyberLLMInstruct [13] and techniques such as RAG help reduce hallucinations by grounding outputs in reliable data. Continual learning methods for efficient updates and LLM verification strategies remain critical priorities [20]. Equally, Human–AI collaboration research must explore trust calibration. Developing AI reasoning interfaces, e.g., saliency maps or threat confidence scores, can improve analyst trust [44, 57]. Insights from medicine and defense could inspire SOC-specific adaptations. Incorporating personalized trust models and social science frameworks can strengthen team dynamics. A tiered autonomy scale (Levels 0–4) could dynamically adjust AI control based on incident severity, workload, and model confidence. Reinforcement learning and Markov decision processes may further optimize task allocation [11, 17, 56]. Embedding autonomy scaling into governance standards like the NIST AI Risk Management Framework is essential for oversight. Controls such as kill switches, audit logging, and fallback protocols ensure accountability. As Agentic AI advances toward autonomous cybersecurity roles, research must evaluate its integration under human oversight in areas like penetration testing and network hardening [25]. Ethical and technical safeguards must ensure such systems align with security goals. Open-source platforms like AutoGen enable prototyping agentic SOCs, which should be evaluated for operational efficiency, safety, and governance. Understanding agentic AI's transition from assistants to autonomous partners is vital for future-ready SOC architectures.